\documentclass[letterpaper, 10 pt, conference]{ieeeconf}

\IEEEoverridecommandlockouts
\usepackage[utf8]{inputenc}
\usepackage[english]{babel}

\usepackage{xspace}
\usepackage{graphicx}
\usepackage{wrapfig}
\usepackage{xspace}
\usepackage{dsfont}

\usepackage{xcolor}
\definecolor{gk}{RGB}{120, 120, 120}
\definecolor{gg}{HTML}{5f9411}
\definecolor{gb}{HTML}{417598}
\definecolor{gr}{HTML}{d15120}
\definecolor{gy}{HTML}{d2ad00}

\usepackage{mathtools}

\makeatletter

\renewcommand\paragraph{\@startsection{paragraph}{4}{\z@}{0.2em}{0em}{\hspace{-0.5em}\noindent\bfseries\normalsize}}
\makeatother

\usepackage{amsmath}
\usepackage{amssymb}
\usepackage{amsthm}

\DeclareMathOperator*{\argmin}{arg\,min}

\usepackage[linesnumbered,ruled]{algorithm2e}
\SetArgSty{textnormal}

\usepackage{listings}
\usepackage{lipsum}

\newcommand{\mydots}{.\hspace{-.1em}.\hspace{-.1em}.}

\usepackage[
activate   = {true},
protrusion = false,
expansion  = true,
kerning    = true,
spacing    = true,
tracking   = false,
auto       = true,
selected   = true,
factor     = 2000,
stretch    = 20,
shrink     = 20,
]{microtype}

\usepackage{csquotes}
\usepackage{cite}

\makeatletter
\let\NAT@parse\undefined
\makeatother
\usepackage[pdfa,colorlinks,bookmarksopen,bookmarksnumbered,allcolors=gr]{hyperref}

\usepackage[nameinlink,capitalise]{cleveref}
\crefname{line}{line}{lines}
\crefname{figure}{Fig.}{Figs.}
\Crefname{figure}{Fig.}{Figs.}
\crefname{equation}{Eq.}{Eqs.}
\Crefname{equation}{Eq.}{Eqs.}
\crefname{section}{Sec.}{Secs.}
\Crefname{section}{Sec.}{Secs.}
\crefname{definition}{Def.}{Defs.}
\Crefname{definition}{Def.}{Defs.}
\crefname{algorithm}{Alg.}{Algs.}
\Crefname{algorithm}{Alg.}{Algs.}
\crefname{assumption}{Asm.}{Asms.}
\Crefname{assumption}{Asm.}{Asms.}

\newcommand{\mf}[1]{\mbox{\cref{#1}}\xspace}

\usepackage{flushend}

\usepackage{soul}
\newcommand{\cameraready}[1]{{{#1}}}

 \newtheorem{definition}{\bf Definition}

\newcommand{\workspace}{\mathcal{W}}

\newcounter{model}

\newcommand{\cmath}[1]{\ensuremath{#1}\xspace}

\newcommand{\dims}{\cmath{d}}
\newcommand{\state}{\cmath{x}}
\newcommand{\start}{\cmath{x_0}}
\newcommand{\goal}{\cmath{\cspace_{\mathrm goal}}}
\newcommand{\cspace}{\cmath{\mathcal{X}}}
\newcommand{\cfree}{\cmath{\cspace_{{\rm free}}}}
\newcommand{\cobs}{\cmath{\cspace_{{\rm obs}}}}

\newcommand{\traj}{\cmath{\pi}}
\newcommand{\trajspace}{\cmath{\Pi}}
\newcommand{\opttraj}{\cmath{\pi^{*}}}

\newcommand{\cost}{\cmath{\mathbf{c}}}

\newcommand{\violationfunc}{\cmath{\mathbf{v}}}

\newcommand{\privacy}{\cmath{\mathbf{p}}}

\newcommand{\privacycost}{\cmath{\cost_{\privacy}}}
\newcommand{\privacyviolation}{\cmath{\violationfunc_{\privacy}}}
\newcommand{\regions}{\cmath{\mathcal{O}}}
\newcommand{\region}{\cmath{{o}}}
\newcommand{\privacyregion}{\cmath{\region_\privacy}}
\newcommand{\privacyregions}{\cmath{\regions_\privacy}}
\newcommand{\privacyweight}{\cmath{\mathbf{w}}}

\newcommand{\true}{\ensuremath{\mathbf{1}}}
\newcommand{\false}{\ensuremath{\mathbf{0}}}

\newcommand{\robot}{\cmath{\mathbf{A}}}

\newcommand{\resub}[1]{{#1}}

\title{\LARGE \bf Robots as AI Double Agents: Privacy in Motion Planning}
\author{Rahul Shome, Zachary Kingston, and Lydia E. Kavraki\thanks{\resub{The authors are affiliated to the Department of Computer Science, Rice University, USA. RS is now affiliated to the School of Computing at the Australian National University, Canberra. \texttt{rahul.shome@anu.edu.au, \{zak, kavraki\}@rice.edu}. This work was supported in part by NSF 1718478, NSF 2008720, and Rice University Funds.}}}

\begin{document}

\maketitle
\thispagestyle{empty}
\pagestyle{empty}

\begin{abstract}
Robotics and automation are poised to change the landscape of home and work in the near future.
Robots are adept at deliberately moving, sensing, and interacting with their environments. 
The pervasive use of robotics promises societal and economic payoffs due to its capabilities---conversely, the capabilities of robots to move within and sense the world around them is susceptible to abuse.  
Robots, unlike typical sensors, are inherently autonomous, active, and deliberate.  
Such automated agents can become \emph{AI double agents} liable to violate the privacy of coworkers, privileged spaces, and other stakeholders.   
In this work we highlight the understudied and inevitable threats to privacy that can be posed by the autonomous, deliberate motions and sensing of robots.   
We frame the problem within broader sociotechnological questions \resub{alongside a comprehensive review}.  
The privacy-aware motion planning problem is formulated in terms of cost functions that can be modified to induce privacy-aware behavior: preserving, agnostic, or violating.     
\resub{Simulated case studies in manipulation and navigation, with altered cost functions, are used to demonstrate how privacy-violating threats can be easily injected, sometimes with only small changes in performance (solution path lengths). Such functionality is already widely available. This preliminary work is meant to lay the foundations for near-future, holistic, interdisciplinary investigations that can address questions surrounding privacy in intelligent robotic behaviors determined by planning algorithms.}
 \end{abstract}

\section{Introduction}
\label{sec:introduction}

Recent advances have introduced robots---particularly robots with manipulation capabilities---into applications such as home assistance~\cite{wirtz2018brave}, healthcare~\cite{Kangasniemi2019}, service~\cite{belanche2020service}, and industry~\cite{bahrin2016industry}.
These settings require the robot to (i) \emph{adapt} to sensed information (e.g., camera images) which is probabilistic and uncertain and (ii) share space and interact with humans, introducing ethical concerns.
We contend that the powerful capabilities of these systems urgently burden us with novel ethical concerns relating to unprecedented use of these systems which, if not addressed now, will lead to dystopian uses of robotics by na\"ive or malicious actors.
Robots are inherently tools of surveillance, have unprecedented access to spaces~\cite{calo202012}, are trusted in ways cameras and other technology is not~\cite{caine2012effect}, and have sensing capabilities that are poorly understood~\cite{lee2011understanding}---a robot that avoids you is also tracking you.
Modern uses of robots combine consumer hardware with intricate frameworks of open-source libraries, middleware~\cite{quigley2009ros}, learned models, and probabilistic algorithms---all of which exacerbate the opacity of a robotic system. Potential abuses are understudied, under-litigated~\cite{hartzog2014unfair} and traditional mitigation strategies are hard or impossible to apply, motivating an urgent need for understanding such threats.

\begin{figure}
    \centering
\includegraphics[width=0.95\linewidth]{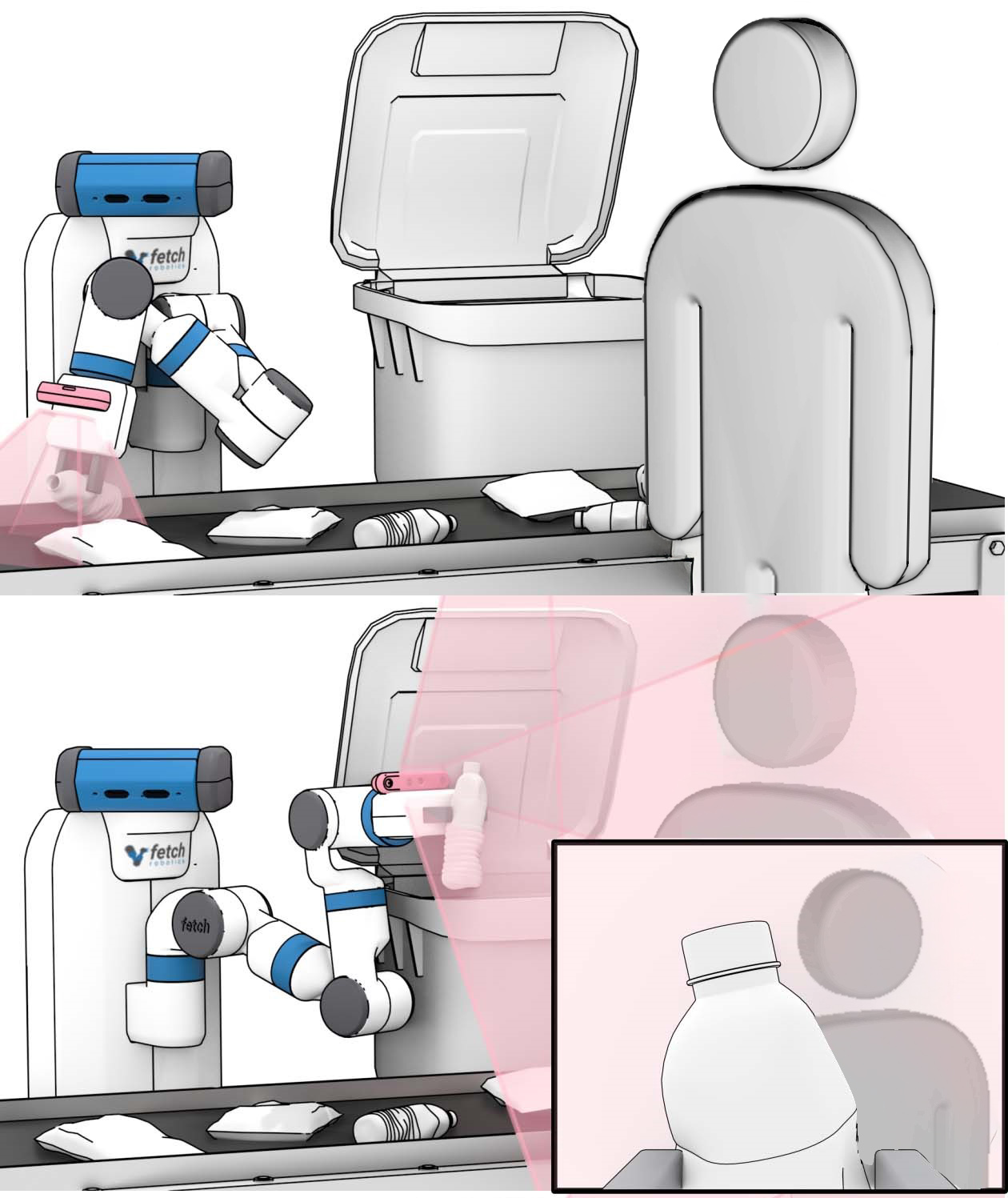}
    \caption{Deployments of robots with sensors can expose threats to privacy from the ability of robots to autonomously use motion planning for choosing how its sensors gather data. In the figure, a sensor attached to a manipulator can gather data about coworkers during typical operation.}
    \label{fig:privacy_planning}
\end{figure}
Privacy violations by robots can take many forms, such as
data over-collection beyond what is strictly necessary for operation that can be used for later inference (as is often observed in web technologies~\cite{datta2014automated,greengard2012advertising}).
Compromised robotic systems with poor security can leak information to unknown third parties \cite{denning2009spotlight, demarinis2019}.
Exfiltration of sensor data or post-hoc analysis of camera data can violate privacy 
by monitoring coworkers or users, extracting privileged information from the workplace and the home \resub{(e.g.,~\cite{roombamit})}.
It behooves us to be wary of the uneasy parallels between the proliferation of robots for generating economic value and surveillance capitalism~\cite{zuboff2018age}.

Focus in the literature has primarily addressed privacy concerns raised by robots and smart devices from a computer vision perspective~\cite{jana2013scanner,kim2019privacy,butler2015privacy,martin2008privacy}. 
There is little work from a privacy standpoint on what makes robots unique---\emph{their ability to move and interact with the physical world.}
In this work we focus on a fundamental element of robotic autonomy---motion planning---and its relation to privacy. 
Robot operators can use custom costs, constraints, and objectives that may place humans co-working with robots in situations that may violate their privacy. 
Abuses to privacy gives rise to what we call \emph{robots as AI double agents}. \resub{To the best of the authors' knowledge, a comprehensive study of privacy implications of generalized motion planning is sorely under-explored.}

\resub{
The key contributions of this work are to present (a) a detailed motivating review of interdisciplinary literature that explains considerations of privacy at the confluence of society, technology, and engineering drawing out the connection to the robotic motion planning problem; (b) a formalization of the privacy in the motion planning problem that identifies privacy-violating functional definitions of feasibility and costs as potential vulnerabilities; (c) a set of motivating case studies based on typical manipulation and navigation scenarios using simple simulations and a straightforward weighted cost function to demonstrate that (i) simple cost function alterations can cause severe privacy-violating behavior, and (ii) privacy-violating behavior can be accompanied by only minor changes in traditional performance metrics (path length).
The technical choices in the simulated study are simple modifications to the motion planning problem, using readily available open-source functionality, that serve to provide an illustrative testbed.
The takeaways from this work points out the clear and present dangers to privacy posed in robotic motion planning.
} 
\section{The Bigger Picture of Privacy and Robotics}
\label{sec:society}

We first take a step back and look at where robotics lies within the broader context of engineering and cyber-physical systems. Many of the privacy considerations attributed to traditional uses and abuses of technology are aggravated by the power of robotic systems to not only be passive sensors, but also be autonomous in the physical space.

\paragraph{Privacy} We must concretely define what we mean by \emph{privacy}~\cite{privacy}. A precise definition is closely tied to societal and legal interpretations in different parts of the world. We choose to refer to GDPR, a push towards common law privacy safeguards~\cite{voigt2017eu}. A closely related definition~\cite{de1981convention} 
promises safeguards that
\emph{``protects the individual against abuses which may accompany the collection and processing of personal data and which seeks to regulate at the same time the transfrontier flow of personal data.''}

Note that we are separating security concerns from those of privacy (e.g., see \cite{dieber2016application,mcclean2013preliminary,dieber2017security,white2016sros,demarinis2019} for ROS and security).
However, privacy violating systems are more readily exploited when security has been compromised.

\paragraph{Engineering Ethics} In the context of robotics, a significant portion of the system design falls on automation deployers, consultants, and engineers. This draws a close connection between the questions of ethical automation and engineering ethics~\cite{nichols1990difficult}. Ethics has been studied as an important aspect of engineering problems where solutions have to trade off ethical considerations and risks versus profit, efficiency, and output.  
The choices made by engineers can have critical societal impacts and unethical choices can have significant fallout. Engineering ethics is also deeply connected to morality and responsibility~\cite{van2012engineering}.    
Intelligent automation lies under the shadow of this complicated relationship between technology, ethics, and society. Beyond sharing many common problems~\cite{lutz2015robocode}, the powerful capabilities of robotics presents unique challenges and threats. 

\paragraph{Cyber-physical Systems} While analysis on privacy has been done in traditional cyber-physical systems~\cite{hassan2019differential,giraldo2017security}, the internet of things~\cite{weber2010internet}, and so on, intelligent robot systems have received little analysis.
There also has been little understanding and observation from policy makers to the threat that robotics bring to privacy, particularly as these systems become more ubiquitous~\cite{hartzog2014unfair,kaminski2016averting}.
There is an uncomfortable relationship between smart home devices and considerations of privacy and legalese~\cite{mills2021alexa, harrigan2017privacy}, including innocuous products like smart toys~\cite{mcreynolds2017toys}.

\paragraph{Robots as Sensors} From the perspective of a robot as a passive sensor platform, there has been much work in preserving privacy~\cite{das2017assisting}, through methods such as obfuscating sensor input~\cite{jana2013scanner}, using degraded images (e.g., anonymizing faces~\cite{kim2019privacy}, reducing quality of the camera feed to a teleoperator~\cite{butler2015privacy}), and redacting relevant parts of the scene~\cite{martin2008privacy}.
There has also been use of ``privacy markers'' that indicate regions that should be removed or redacted from sensors~\cite{schiff2009respectful, raval2014markit}, automatically detecting these regions~\cite{fernandes2016detection}, and also extended to a case with mobile robots~\cite{rueben2016evaluation}.
However, all these methods merely operate on the camera feed passively, and do not actively direct what information the sensor should gather.
Even with minimal data, powerful inference can identify individuals (e.g., with motion~\cite{loula2005recognizing}). In general, robots must collect only the data they need~\cite{kaminski2016averting}.

\paragraph{Robots around Humans}
Trust is essential for embodied systems to operate reliably near humans~\cite{glikson2020human}.
Moreover, people are more ``comfortable'' with robots rather than unembodied cameras, and more willingly expose themselves to privacy violations~\cite{caine2012effect}, and misunderstand the full capabilities of robotic systems to gather information~\cite{lee2011understanding, rueben2020estimating}. There are certain qualities that humans expect from robots, and how that relates to how robots can ``fly under the radar'' when doing things. This relates to intention-aware planning~\cite{glikson2020human}.

\paragraph{Privacy and Learning} There has been much recent work on using machine learning-based methods for robot control, particularly in learning from human demonstrations~\cite{argall2009survey,ravichandar2020recent}.
Learning based methods require large amounts of data, which is at odds with privacy concerns that require minimal data collection.
There has been work in addressing privacy in deep learning~\cite{shokri2015privacy,al2019privacy}, namely in differential privacy~\cite{dwork2008differential,abadi2016deep}, but little from this literature has been applied to robotics and control~\cite{han2018privacy}, particularly in the context of manipulation.

\resub{Given the complexity of machine learning-based models, there is also potential for subterfuge, e.g., adding undetectable backdoors to modify behavior of a system~\cite{goldwasser2022planting}. Such backdoors could be used to induce malicious behavior in models used to nominally preserve privacy without the awareness of stakeholders.
}

\begin{figure}
    \centering
\includegraphics[width=0.8\linewidth]{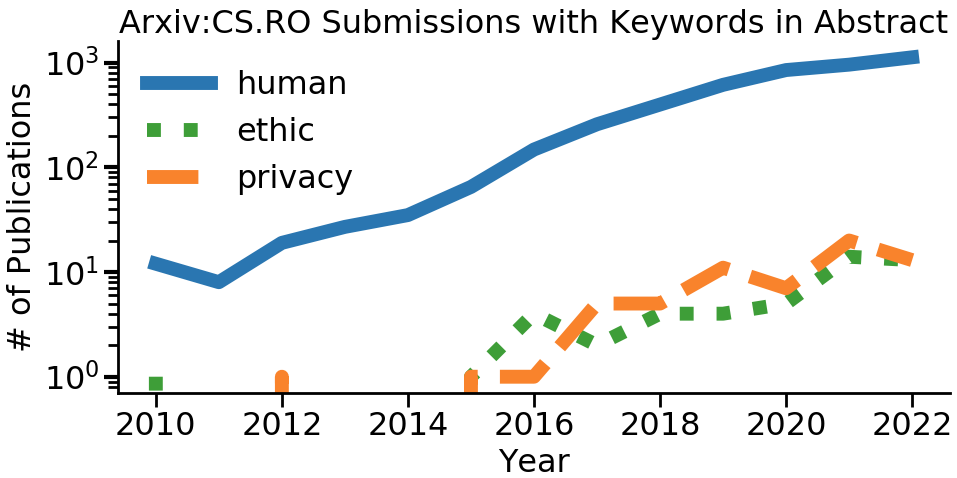}
    \vspace{-0.1in}
    \caption{The figure shows the logarithmic growth of submissions to \emph{Arxiv:CS.RO} with the keywords \emph{human}, \emph{ethic}, and \emph{privacy} in their abstract between 2010 and \cameraready{2022}.
    Privacy and ethical concerns are far outpaced by applications that interact with humans.
    }
    \label{fig:arxiv}
\end{figure}

\paragraph{Privacy in Robotics Research}
Privacy is pressing issue throughout the broader AI community~\cite{jobin2019global}.
Robotics in particular is a concern for privacy, given the direct nature of a robotic system as a tool of surveillance, with sensing an inherent part of a robot's ability to understand the world~\cite{rueben2018themes}.
However, considerations of privacy in the field of robotics have been far outpaced by the potential usages that might pose threat.
\mf{fig:arxiv} is an approximation of what is an undeniable trend---robotics solutions and applications that have the potential to interface with humans are growing beyond the understanding of privacy (and ethics) in these contexts. 
\cite{eick2020enhancing} makes a similar observation.

There is some work in understanding privacy concerns for social robots~\cite{lutz2019privacy,fosch2020gathering}---that is, systems primarily designed for human interaction and entertainment.
Systems, such as household robots, have been shown to have poor security and privacy properties~\cite{denning2009spotlight}.
We focus instead on general scenarios where the intended purpose, such as a logistics task, can be compromised by motion planning.

\paragraph{Privacy and Planning}
There has been prior work in incorporating deliberate reasoning about sensing within planning. 
Generally falling into the category of ``active vision'' \cite{chen2011active}, work exists in autonomous surveillance \cite{abidi2009survey}, searching for objects \cite{shubina2010visual, adu2021probabilistic}, and chronicling~\cite{rahmani2020planning,shell2019planning}.
Some work has used the capabilities of robotic arms~\cite{rakita2018autonomous} in assistive applications~\cite{armbrust2011using}. There has been work for protecting the privacy of robots themselves~\cite{tsiamis2019motion,li2019coordinated,zheng2020adversarial}, but these works typically apply to mobile or aerial applications.
Drones form a common platform of choice for visibility-aware problems~\cite{nieuwenhuisen2019search} with some work on aspects of privacy~\cite{9197564,pan2019unmanned,kim2019udipp,mcbride2009beyond,finn2012unmanned,blank2018privacy}.
These works focus on clearly defined areas to avoid (e.g., similar to explicitly marked privacy zones) but only deal with low-dimensional systems (e.g., mobile robots and drones). General robots with many joints---like manipulators that are beginning to be more broadly deployed---pose significant challenges due to the high-dimensionality of their search space.  
Addressing the problem for general platforms and  manipulators is necessary to apply to broader scenarios.

\section{Privacy-Aware Motion Planning}
\label{sec:technical}

In this section we take a closer look at the role privacy plays in a fundamental aspect of robotics---motion planning. We introduce some of the technical and theoretical tools necessary to define and understand elements of privacy in motion planning. We focus on a threat model where a robot fitted with a sensor collects data while moving. Privacy-aware motion planning will be defined in terms of data collected on privacy-sensitive regions. The vulnerabilities framed in this section can potentially lie exposed to deployers and end-users. Modifications to parameters, like cost functions, in motion planning can lead to altered behavior by robots acting as AI double agents.

\begin{definition}[AI Double Agent]
    An AI double agent is a robot, that by virtue of altered reasoning and planning, exhibits autonomous behavior which violates the privacy of any human agent in the robot's workspace.
\end{definition}

\subsection{Privacy-Aware Motion Planning}

A robotic agent \robot with \dims degrees of freedom, e.g., robotic arm with \dims joints, is situated in a workspace $\workspace\in\mathbb{R}^3$ with obstacle regions $\region\subseteq\workspace$.
The robot has a \dims-dimensional configuration space  $\cspace\subseteq\mathbb{R}^d$.
Each configuration of the robot can be checked for feasibility, typically defined in terms of being collision-free with the obstacles. 
Denote a boolean feasibility function as $\violationfunc: \cspace\rightarrow\{\true,\false\}$. 
The invalid subset corresponds to $\cobs=\{\state \mid \ \violationfunc(\state)=\false, \state\in\cspace\}$, while the valid subset is $\cfree=\cspace\setminus\cobs$. 
A motion planning problem requires connecting a start configuration $\start\in\cspace$ to a goal region \goal with a continuous, collision-free, time-parameterized trajectory $\traj: [0,1]\rightarrow\cfree$ where $\traj[0]=\start, \traj[1]\in\goal$. 
Given all possible such feasible trajectories $\trajspace \ni \traj$, a cost function $\cost: \trajspace \rightarrow \mathbb{R}_{\geq 0}$ assigns a non-negative real number to a trajectory. The cost is typically considered to be (or some function proportional to) the Euclidean path length. An optimal solution corresponds to the minimum cost $\opttraj \in \argmin_{\traj\in\trajspace}\ \cost(\traj)$.

In this work we introduce the element of privacy into the motion planning problem. For notational clarity we will use subscripts with $\privacy$ to denote privacy-aware variants. 
A privacy-sensitive regions is a region of the workspace which is associated with requirements for privacy preservation is denoted by $\privacyregion\subseteq\workspace$. A set of $k$ such regions is denoted by $\privacyregions = \{\privacyregion^1\cdots\privacyregion^k\}$.
A privacy feasibility function applies to a configuration and is denoted by $\privacyviolation: \cspace \rightarrow \{\true,\false\}$. 
Without loss of generality we will consider privacy-violating evaluations when $\privacyviolation$ evaluates to $\false$.
A non-negative privacy-aware cost is defined along a trajectory $\privacycost:  \trajspace \rightarrow \mathbb{R}_{\geq 0}$.

\resub{
The evaluation of both the constraint and cost $\privacyviolation$ and $\privacycost$ will depend upon the privacy regions $\privacyregions$. The exact nature of this relationship will be affected by the precise setting under consideration including the kinematics of the robot, the attachment of the sensor, the sensing model (for instance visibility cone for a camera, etc). Our general formulation will leave these as necessary privacy-aware pieces within the otherwise typical motion planning problem.
}
Note that the definition of the problem thus far can be applied to many general combinations of robots, sensors, and privacy regions. 

\subsection{Types of Privacy-Awareness}
The definitions of $\privacyviolation$ and $\privacycost$ can allow interactions with privacy regions $\privacyregions$ with three types of privacy-awareness:

\noindent\textbf{Privacy-Agnostic \resub{($\violationfunc,\cost$)}} classical motion planning has unmodified feasibility and cost functions.

\noindent\textbf{Privacy-Preserving \resub{($\privacyviolation^+,\privacycost^+$)}} choices penalize privacy violations along robot motions.

\noindent\textbf{Privacy-Violating \resub{($\privacyviolation^-,\privacycost^-$)}} choices promote privacy violations along robot motions.

\resub{
Privacy-aware behaviors present a choice of functional alternatives. This leads us to the step where these are defined, which is up to the problem designer or deployer.

\begin{definition}[Motion Planning Double Agent]
The design of privacy-violating $\privacyviolation^-$ or $\privacycost^-$ 
    to replace the privacy-agnostic feasibility and cost functions, creates a double agent generating motion plans for the privacy-violating variant of the problem.
\end{definition}
}

\subsection{Privacy as a Secondary Objective}
The privacy-aware cost function $\privacycost$, which is either privacy-preserving or violating behavior, can encode or be a part of multiple objectives~\cite{stewart1991multiobjective} within the motion planning problem. The threat model of interest, and indeed of greater risk and harder to detect, is expected to involve robots that perform their primary automation operation satisfactorily while \emph{also} achieving a secondary privacy-aware objective. 
For instance, consider a possible combination of the length of solution path (a traditional cost in motion planning) and the privacy region visibility by a camera attached to the robot. The manner of this combination 
can lead to different flavors of pareto-optimal~\cite{goldin2021approximate} problems,
while the continuous nature of the problem can relate to cost maps~\cite{devaurs2013enhancing}.

\emph{Key to the broader scope is not the prescription of such a specific cost and feasibility. Rather, in the next section, we show that it suffices to design \textbf{minor changes to the cost function in standard motion planners to make them privacy-aware}. This is particularly interesting because of the relatively low barrier of access, as it might be possible for deployers or end-users with enough expertise to modify the parameters and modules of motion planning.} 
\section{Motivating Simulated Case Study}
\label{sec:results}

In this section we demonstrate how---using a candidate modified cost function---privacy-aware behavior can be injected into normal operation of a motion planner.

\subsection{Candidate Model of Privacy-Aware Cost Function}
Consider a PRM*~\cite{karaman2011sampling} as the motion planner that reports shortest paths over roadmaps~\cite{kavraki-svestka1996probabilistic-roadmaps-for} constructed in the robot's configuration space. 
\textit{Custom cost functions} change the graph edge weights, altering the discovered solutions executed by the robot.
This basic functionality is readily available through powerful open-source libraries~\cite{sucan2012open}. 
\resub{While being careful not to prescribe what a privacy-aware cost function should be, we provide a straightforward candidate for studying the effects introduced by weighted modifications to the Euclidean path length cost function.} 
The trajectory $\traj_\privacy$ is weighted multiplicatively or fractionally using a \emph{privacy weight} ($\privacyweight$) parameter (such that $|\privacyweight|\geq 1$) depending upon the interaction with the privacy regions. A negative weight is privacy-violating. The total cost will be calculated over a discretization ($\Delta \traj$) of the trajectory $\traj_{\privacy}$. $|\privacyweight|=1$ is privacy-agnostic.

\begin{figure*}[ht!]
    \centering
    \includegraphics[trim={0 0 0 1.5cm},clip, width=0.3\linewidth]{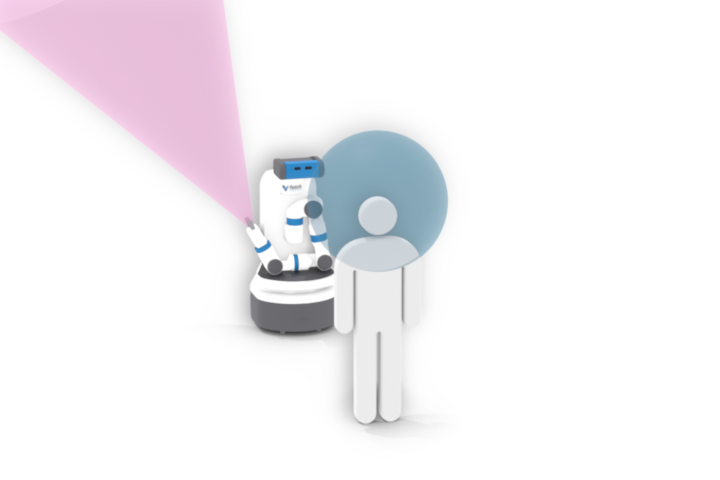}
    \includegraphics[trim={0 0 0 1.5cm},clip, width=0.3\linewidth]{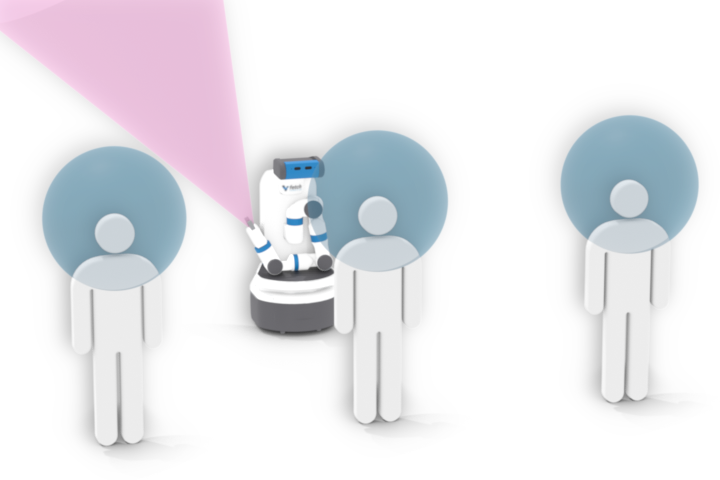}
    \includegraphics[trim={0 0 0 1.5cm},clip, width=0.3\linewidth]{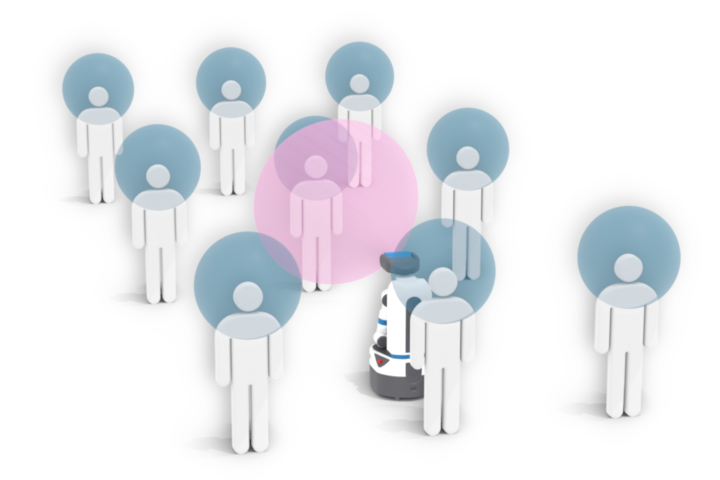}\\
    \includegraphics[trim={0 0 0 0},clip, width=0.3\linewidth]{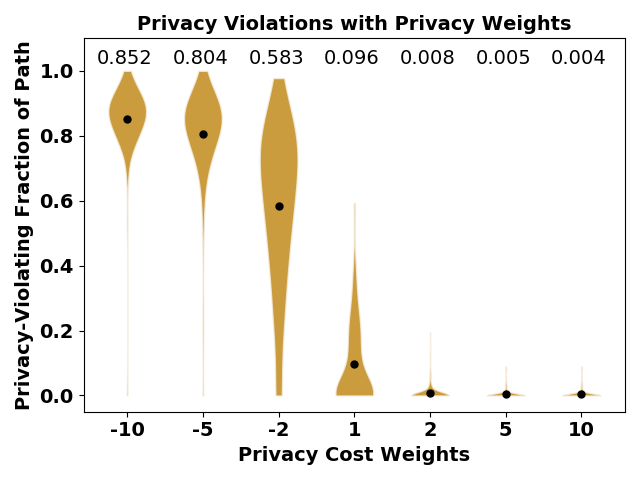}
    \includegraphics[trim={0 0 0 0},clip, width=0.3\linewidth]{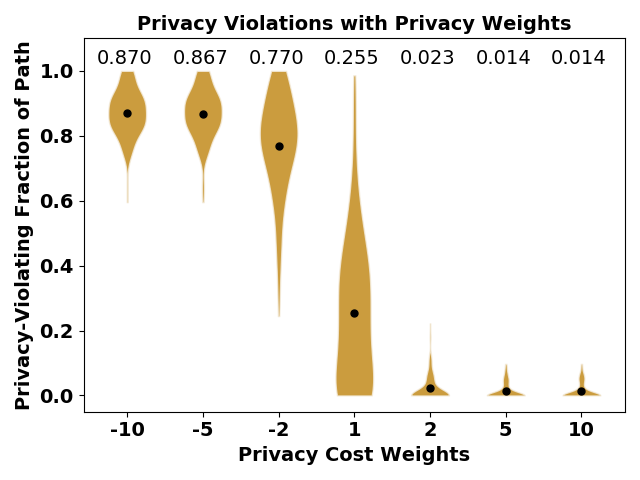}
    \includegraphics[trim={0 0 0 0},clip, width=0.3\linewidth]{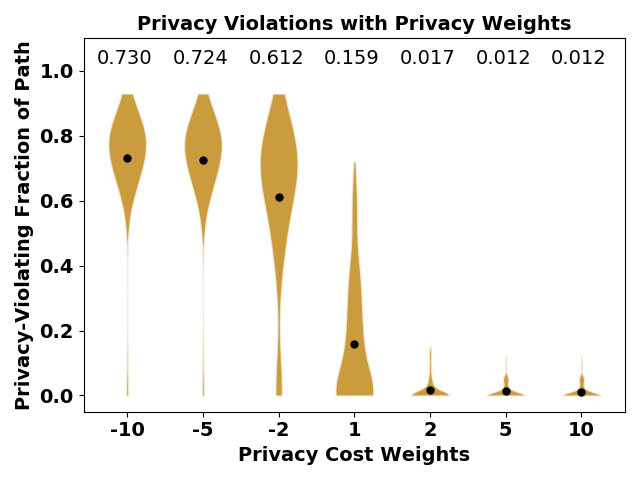}
    \includegraphics[trim={0 0 0 0},clip, width=0.3\linewidth]{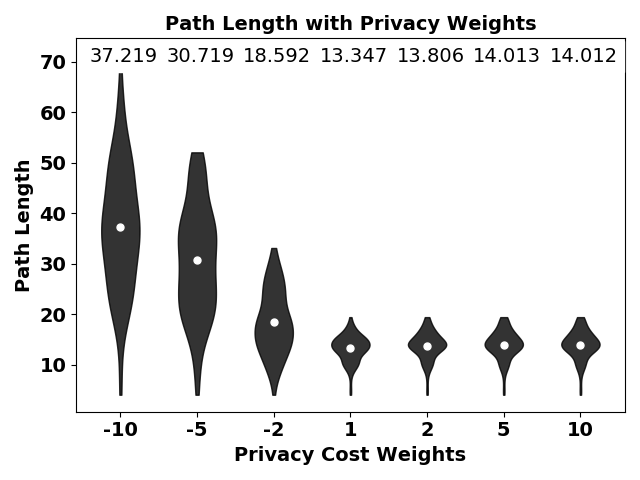}
    \includegraphics[trim={0 0 0 0},clip, width=0.3\linewidth]{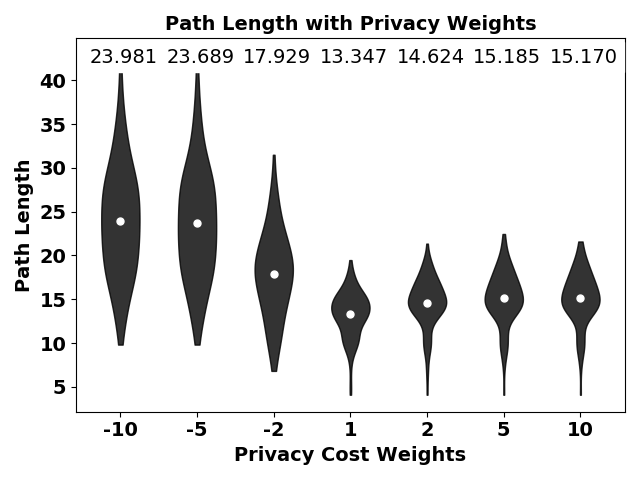}
    \includegraphics[trim={0 0 0 0},clip, width=0.3\linewidth]{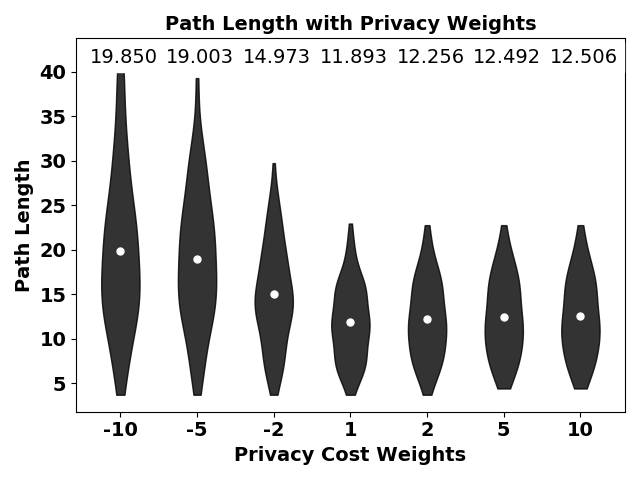}\\
    \includegraphics[trim={0 0 0 0},clip, width=0.3\linewidth]{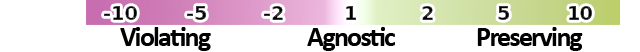}
    \includegraphics[trim={0 0 0 0},clip, width=0.3\linewidth]{figures/footer.png}
    \includegraphics[trim={0 0 0 0},clip, width=0.3\linewidth]{figures/footer.png}
    \caption{(Top) A Fetch robot (with camera visibility cone in pink) controlling either the \textbf{arm+torso} with a hand camera (left, middle) or the \textbf{base+head} with a head camera (right). (From left to right) 1, 3, and 9 privacy regions (blue spheres) overlaid on humans in the workspace. Violin plots (with mean markers) for 100 random runs with the privacy violating fraction (middle) and path lengths (bottom) for different privacy weights. Mean values are given at the top. }
    \label{fig:bench}
\end{figure*}

\paragraph{Privacy-Preserving ($\privacyweight>1$)} 
\begin{equation}
    \privacycost^+(\traj_{\privacy}) = \sum_{\Delta \traj \in \traj_{\privacy} } \begin{cases}
        \privacyweight \| \Delta \traj \| \quad \text{\resub{if privacy violated}}\\
        \frac{1}{ \privacyweight } \| \Delta \traj \| \quad \text{\resub{otherwise}}
    \end{cases}
    \label{eq:privpreseq}
\end{equation}  

\paragraph{Privacy-Violating ($\privacyweight<-1$)} 
\begin{equation}
    \privacycost^-(\traj_{\privacy}) = \sum_{\Delta \traj \in \traj_{\privacy} } \begin{cases}
        \frac{1}{ |\privacyweight| } \| \Delta \traj \| \quad \text{\resub{if privacy violated}}\\
        |\privacyweight| \| \Delta \traj \| \quad \text{\resub{otherwise}}
    \end{cases}
    \label{eq:privvioleq}
\end{equation}

Here $|\cdot|$ and $\|\cdot\|$ denote the absolute value and Euclidean arc length. \resub{In essence, all that is needed is an approximation of privacy violation (for instance, intersection with a camera cone with $\privacyregions$), and functional choices that penalize or promote the privacy preservation or violation. 
Due to the generality of the underlying planning, this should apply to a large variety of sensor-attached high-dimensional robotic platforms. Though the cost function weighting represents a simple alteration, what is not obvious is \emph{how the cost function affects the double agent's privacy and efficiency?} }

\subsection{Case Studies}
\resub{
To highlight motivating scenarios, we focus on two case studies with cameras attached to the robot while it moves. The camera visibility is approximated by a cone (shown in pink in~\mf{fig:bench} top) defined similar to the specifications of an \emph{Intel Realsense} camera (a 42$^\circ$ field of view and 2m range). While sensing these privacy regions poses its own challenges, here we choose to focus on the effects of planning by assuming these as input. Our motivating scenarios will introduce typical workspace settings where specific areas might be expected to contain these privacy-regions.
Humans are represented as static mesh obstacles with spherical approximations (40cm radius) of privacy regions around their heads.

\noindent\textbf{Manipulation: } A manipulator with a camera attached to its wrist is set up in a workspace opposite to human collaborator(s) or customer(s). These types of settings are common to \textit{warehouse automation settings or service industries}.
The task itself is typically concentrated in the shared workspace between the robot and the human, for e.g., a table, a cashier's desk, a counter, etc. The robot is free to move the sensor (wrist) unencumbered, as long as it reaches its planning goal and avoids obstacles. The camera interacts with the regions of the workspace expected to contain human occupancy (spherical approximations shown in~\mf{fig:bench}~\textit{(left, middle)} for one and three individuals respectively).

\noindent\textbf{Navigation: } A mobile robot with a camera attached to a controllable joint is set up to navigate through a planar workfloor filled with human co-workers or crowds. Such scenarios will come up in a variety of \textit{mapping, navigation, cleaning, and monitoring} tasks where the mobile robot is operating within the floorplan while avoiding the obstacles (here humans). Since the head camera is freely controllable during its motions, the camera interacts with the regions of the workspaces expected to contain human occupancy (spherical approximations shown in~\mf{fig:bench}~\textit{(right)} for nine individuals).
}

\cameraready{\textit{Experimental Details: }} 
The simulation uses a Fetch in two controllable modes: a) \textbf{arm+torso}: a camera attached to its wrist (8-dim \cspace) and b) \textbf{base+head}: a controllable head camera (5-dim \cspace). 
A PRM*~\cite{karaman2011sampling} is constructed 
and reused across the experiments. Each choice of privacy cost alters the weights on this roadmap.
Given random valid starts and goals, uniform cost search reports a solution.
The metrics reported for the resultant paths are a) \textbf{privacy violation fraction}, which is the fraction of the path where the sensing cone intersects with any of the privacy regions, and b) \textbf{path length}. 
$\privacyweight$ (\mf{eq:privpreseq,eq:privvioleq}) is chosen from $\{1,\pm2,\pm5,\pm10\}$. The code was implemented with Robowflex~\cite{robowflex} and OMPL~\cite{sucan2012open}.

\subsection{\resub{Effects of Changing Privacy Cost ($\privacycost$)}}
\label{sec:results:sim}

\begin{figure*}[t]
    \centering
    \includegraphics[trim={0 0cm 0 0cm},clip, width=0.32\linewidth]{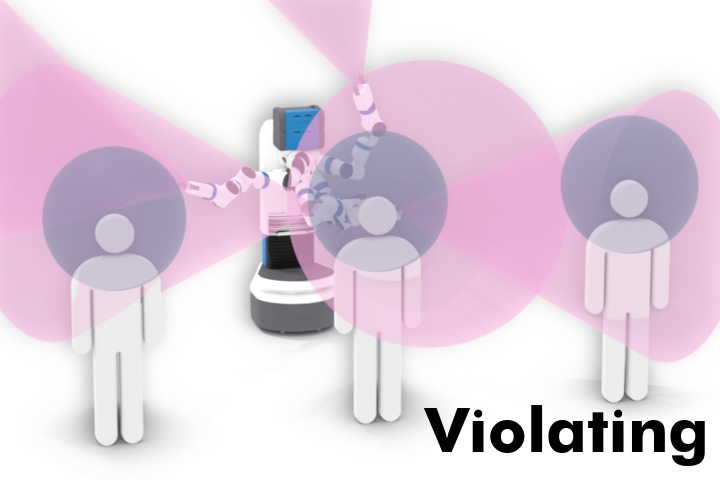}
    \includegraphics[trim={0 0cm 0 0cm},clip, width=0.32\linewidth]{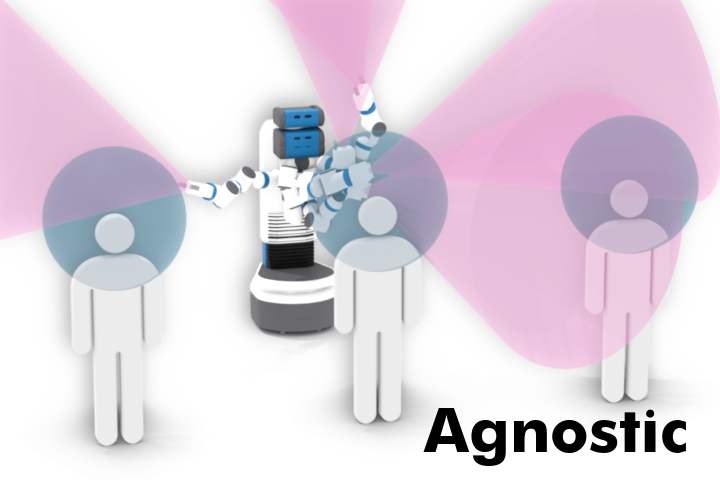}
    \includegraphics[trim={0 0cm 0 0cm},clip, width=0.32\linewidth]{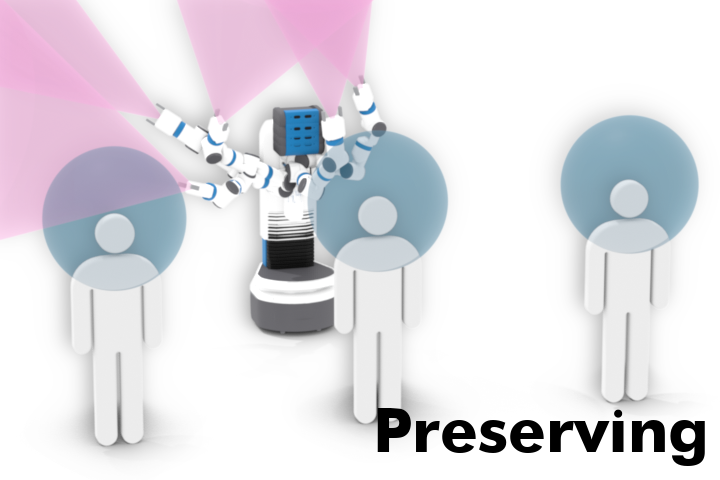}
\caption{The visualization of different privacy-aware behavior: violating (left),  agnostic (middle), and preserving (right) showing snapshots of the pink camera cone along motions solving the same problem. Even the agnostic path sweeps the camera cone over the right privacy region. The violating path makes the camera cone intersect with all the privacy regions while the preserving one avoids them altogether.}
    \label{fig:paths}
\end{figure*}

\mf{fig:bench} demonstrates the different behaviors obtained by changing $\privacycost$ by tuning $\privacyweight$. All the motions successfully connect the start and goal but differ in the interactions between the camera cone and privacy regions (\mf{fig:bench} middle). Privacy-violating (negative) weights significantly increase the portion of the motions in which the camera \textit{lingers} on the privacy regions. The privacy-violating solutions are also longer (\mf{fig:bench} bottom) further increasing the data gathered. Note how an example violating motion (\mf{fig:paths} left) ends up gathering data on all the regions. In contrast, the privacy-agnostic paths are shorter but are still liable to capture data on the privacy regions. The 3-region benchmark shows violations along a quarter of the motions on average (\mf{fig:bench} middle-middle). This motivates explicit reasoning about privacy preservation. Privacy-preserving paths immediately show the benefits of  penalizing intersections between the cone and the privacy regions, with such violations dropping below 0.25\% in all cases, while still maintaining relatively short paths. Note the privacy-preserving motion in \mf{fig:paths} (right) completely avoids the privacy regions. Increasing the absolute magnitude of $\privacyweight$ strengthens the preserving or violating behavior.

\resub{A trade-off arises in the data between privacy violation and performance, both of which might be attributed economic value in the design of automation. Large negatively weighted privacy violations are associated with large path length increases. Interestingly, in certain cases ($\privacyweight=-2$) high privacy violations ($>0.58$) arise on average with only a small dip in performance ($<1.4$ times the agnostic path length) in all the case studies. We highlight that double agent might be deliberately designed to \textbf{aggravate low-privacy, high-performance behaviors}.
}
Additionally, trade-offs can arise when such data gathering might be necessary for detecting humans for safe operation \resub{and handling cases of dynamic or uncertain detections}. 
Multiple sensors and targeted data gathering among multiple regions also poses risks.
 \resub{Though the studied cost function is \emph{not} prescriptive and several alternate models exist, the chief takeaways remain unchanged.}

\resub{
\noindent\textbf{Key Causes for Alarm: } The observations from our simple simulated case studies already illustrate serious causes for concern. (a) With relatively straightforward alterations to the cost function, \textbf{drastic privacy-violating behavior was introduced}. (b) Significant privacy-violating behavior \textbf{might only introduce relatively small changes to the performanc}e (privacy-agnostic path length).
(c) \textbf{Functionality to plug in custom cost function is readily available} in open-source planning libraries. (d) Privacy-preserving behavior \textbf{only emerges when deliberately included} in the cost function.

}

\section{Discussion}
\label{sec:discussion}
The work presented here has highlighted the privacy threats exposed by robotic double agents capable of autonomous planning and reasoning to move and sense in deliberate ways. 
\resub{This work contributes a thorough review of the interdisciplinary connections between privacy and motion planning, formulates the role of privacy in the motion planning problem, and showcases imminent privacy-violating threats using motivating simulated case studies and straightforward cost function modifications to motion planning. The work calls for a more consolidated investigation.}

\paragraph{Human-Centric Factors} The human aspect is essential in the problem. 
\cameraready{The current simulated studies} motivate the need for a deeper understanding of how humans perceive privacy-aware robotic behavior. The human-centric factors here are intrinsic to privacy, 
\cameraready{necessitating the investigations to be}
centered around the rights and protections of humans.  
There has been related human-robot interaction work that studies the anthropocentric principles to align robot motion to human expectations in intention-aware planning \cite{bandyopadhyay2013intention, bai2015intention, chen2018planning}, the consideration of  ethical or human-focused value alignment \cite{hadfield2016cooperative}, and planning with legibility or human interpretation as an objective~\cite{lichtenthaler2013towards,dragan2013legibility,dragan2015effects}. There has also been work on communication~\cite{huang2019enabling}, understanding robot motion~\cite{beetz2010generality,dragan2014familiarization}, and parameterized social interactions~\cite{lu2013tuning} in navigation. \cameraready{Tradeoffs can exist between human-awareness in robots and privacy.
The current study also raises questions about the context and expectations of privacy from the robot---an understanding by the human of an ongoing or imminent privacy violation.}

\paragraph{Verification and Mitigation} This work admittedly raises more questions than answers with respect to the ways in which the threats posed by robotic planning can be identified and mitigated. While intentional deployment and recommended practices of privacy-preserving planning can achieve some headway, it does not resolve the issue of the bad actor or even a na\"ive one (who exposes threats that exist even in privacy-agnostic planning). There needs to be broader discussions among social scientists, ethicists, and policy-makers to inform rules, regulations, and deployments. The understanding of the human factors can also inform community stakeholders like coworkers or end-users. Privacy is also connected to vulnerabilities in open-source and general-purpose software. It is critical that the robotics community recognizes these imminent threats to step towards a future that avoids the worst of these threats. 

\paragraph{Call to Action} We look into motion planning as one of the most fundamental capabilities of autonomous robots that can be abused for privacy violations. We demonstrate the imminent threats that exist on the near-horizon using technologies and solutions that exist today. The increased incidence of robots in our work and home will raise uncomfortable questions of how these robots are harvesting data and protecting privacy. 
A call to action is needed for the robotics community to reach out to other stakeholders to build towards a future with useful robots that are both capable and comply with fundamental human values, such as privacy.

\bibliographystyle{IEEEtran}

\end{document}